\documentclass[conference]{IEEEtran}
\usepackage{amsmath}
\usepackage{multirow}
\usepackage{graphicx}
\begin{document}

\title{Multi-Site Clinical Federated Learning using Recursive and Attentive Models and NVFlare}

\author{Won Joon Yun$^{1}$, Samuel Kim$^{2}$, and Joongheon Kim$^{1}$
\\ \\
\IEEEauthorblockA{$^1$~School of Electrical Engineering, Korea University, Seoul, Republic of Korea}
\IEEEauthorblockA{$^2$~Cipherome Inc., San Jose, California, USA}
}

\maketitle

\begin{abstract}
The prodigious growth of digital health data has precipitated a mounting interest in harnessing machine learning methodologies, such as natural language processing (NLP), to scrutinize medical records, clinical notes, and other text-based health information. Although NLP techniques have exhibited substantial potential in augmenting patient care and informing clinical decision-making, data privacy and adherence to regulations persist as critical concerns. Federated learning (FL) emerges as a viable solution, empowering multiple organizations to train machine learning models collaboratively without disseminating raw data. This paper proffers a pragmatic approach to medical NLP by amalgamating FL, NLP models, and the NVFlare framework, developed by NVIDIA. We introduce two exemplary NLP models, the Long-Short Term Memory (LSTM)-based model and Bidirectional Encoder Representations from Transformers (BERT), which have demonstrated exceptional performance in comprehending context and semantics within medical data. This paper encompasses the development of an integrated framework that addresses data privacy and regulatory compliance challenges while maintaining elevated accuracy and performance, incorporating BERT pretraining, and comprehensively substantiating the efficacy of the proposed approach.
\end{abstract}

\begin{IEEEkeywords}
Federated Learning, Clinical Data, Language Model
\end{IEEEkeywords}

\section{Introduction}
The burgeoning expansion of digital health data has incited a mounting interest in employing machine learning methodologies, such as natural language processing (NLP), to scrutinize medical records, clinical notes, and other text-based health information. NLP techniques have exhibited considerable potential in augmenting patient care and informing clinical decision-making~\cite{xu2019hybridalpha}. However, the sensitive nature of health data and the imperative for adherence to regulations, such as the Health Insurance Portability and Accountability Act (HIPAA), pose considerable challenges in terms of data privacy and security. Federated learning (FL) constitutes a distributed machine learning paradigm that empowers multiple organizations to collaboratively train a model without sharing raw data, thereby ensuring data privacy and legal compliance~\cite{yang2019federated,pieee202105park,ICDCS22}. FL facilitates the cooperative training of a shared machine learning model across multiple clinics while safeguarding patient data privacy and complying with regulations. Furthermore, the collaborative training using data from diverse clinics promotes the development of more robust and accurate models, which may potentially generalize better to unseen data, culminating in enhanced diagnostics. Additionally, FL obviates data silos that frequently arise in multi-site clinics, as it allows institutions to learn from one another's data without infringing upon privacy regulations. This collaborative approach can foster increased knowledge sharing and improved patient outcomes. As a result, FL can address real-world data discrepancies, such as varying data quality, data distribution, and data labeling practices across different clinics, rendering the shared model more applicable to heterogeneous clinical settings and populations.

Motivated by the indispensability of FL, this paper proffers a pragmatic approach to medical NLP by amalgamating FL, NLP models, and NVFlare. We introduce two exemplary NLP models: the Long-Short Term Memory (LSTM)-based model and Bidirectional Encoder Representations from Transformers (BERT)~\cite{devlin2018bert}. NVFlare, devised by NVIDIA, is a versatile and scalable framework for FL that delivers system reliability, privacy preservation, and optimal resource allocation~\cite{nvidia2021nvflare}. By integrating these components, the proposed framework tackles the challenges of data privacy and regulatory compliance whilst maintaining elevated accuracy and performance.

The salient contributions are tri-folded: First, this paper integrates the FL framework that addresses the challenges of data privacy and regulatory compliance while maintaining high accuracy and performance in medical NLP; Second, this paper incorporates not only a general training method but also BERT pretraining, which broadens the applicability of the proposed framework; Lastly, this paper comprehensively demonstrates the efficacy of the proposed approach, which transitions from traditional LSTM-based models to BERT, thereby establishing the practicality of the reference framework.

This paper is structured as follows: Sec. \ref{sec:2} delineates the related work encompassing FL, NLP models, and NVFlare; Sec. \ref{sec:3} expounds upon the methodology for integrating these technologies; Sec. \ref{sec:4} deliberates the results and performance of the proposed approach; and Sec. \ref{sec:5} concludes the paper with a discourse on future research directions.

\section{Related Work}\label{sec:2}
\subsection{Federated Learning with Medical NLP Models}\label{sec:2-1}
FL is a distributed machine learning paradigm that allows multiple organizations to collaboratively train a model without sharing raw data~\cite{yang2019federated}. In this approach, each organization trains a local model on its data and submits the model updates to a central server. The server aggregates these updates to produce a global model, which is then disseminated back to the participating organizations for further local updates. This iterative process continues until the desired level of accuracy and convergence is achieved. FL's key advantage lies in preserving data privacy by retaining sensitive information within the boundaries of each participating organization. This approach has garnered significant attention in healthcare, where data privacy concerns and regulatory compliance are of paramount importance~\cite{rieke2020future}.

In the context of medical NLP, FL enables the development of robust models that leverage the expertise of multiple healthcare institutions while adhering to privacy regulations. By combining FL with advanced NLP models, such as LSTM and BERT, it is possible to create powerful solutions for tasks like medical entity extraction, relation extraction, clinical document classification, generating medical reports, and predicting patient outcomes from clinical notes, all while preserving data privacy.

LSTM-based models are often preferred in medical NLP tasks due to their capacity to handle sequential data and capture long-range dependencies~\cite{ICLR2016}. On the other hand, BERT is an advanced language model that has demonstrated exceptional performance in various NLP tasks, surpassing traditional LSTM-based models \cite{devlin2018bert}. BERT is pre-trained on large text corpora and fine-tuned on specific tasks, enabling it to discern intricate relationships and contextual information within the text. BERT's bidirectional nature permits it to apprehend the context of words from both left and right directions, which is particularly advantageous for analyzing complex and domain-specific language found in medical data.

\subsection{Related Federated Learning Frameworks}\label{sec:2-2}
This section presents an overview of the prominent FL frameworks extensively employed in both industrial and academic settings. These frameworks include:\\
\noindent $\bullet$ \textit{TFF} \cite{tensorflow-federated}: TFF, an open-source framework developed by Google, facilitates machine learning and other computations on decentralized data. It enables developers to implement federated learning algorithms using the high-level APIs of TensorFlow.\\
\noindent $\bullet$ \textit{PySyft} \cite{pysyft-github}: The OpenMined community developed PySyft, a Python library providing federated learning, secure multi-party computation, and differential privacy capabilities. PySyft extends widely-used deep learning libraries such as PyTorch and TensorFlow to support secure and privacy-preserving machine learning.\\
\noindent $\bullet$  \textit{FATE} \cite{fate}: WeBank's open-source FL framework, FATE, is designed to facilitate secure computations on distributed data. It encompasses various federated learning algorithms, including vertical and horizontal federated learning.\\
\noindent $\bullet$  \textit{LEAF} \cite{leaf}: LEAF, an open-source benchmarking framework for federated learning, was developed by researchers at Carnegie Mellon University. It offers a collection of datasets, pre-processing tools, and evaluation metrics to enable fair comparisons among different federated learning algorithms.

In comparison to these frameworks, \textit{NVFlare} delivers superior system reliability, privacy preservation, and optimal resource allocation for distributed machine learning~\cite{nvidia2021nvflare}. The framework is designed to be flexible and scalable, promoting collaboration among multiple healthcare institutions in training deep learning models without sharing raw data. While BERT has not been previously implemented on NVFlare, this paper demonstrates the integration of medical NLP models and NVFlare, thereby rendering it an ideal platform for the development of privacy-preserving medical NLP solutions.

\section{Methodology} \label{sec:3}
\begin{figure}[t!]
    \centering
    \includegraphics[width=\columnwidth]{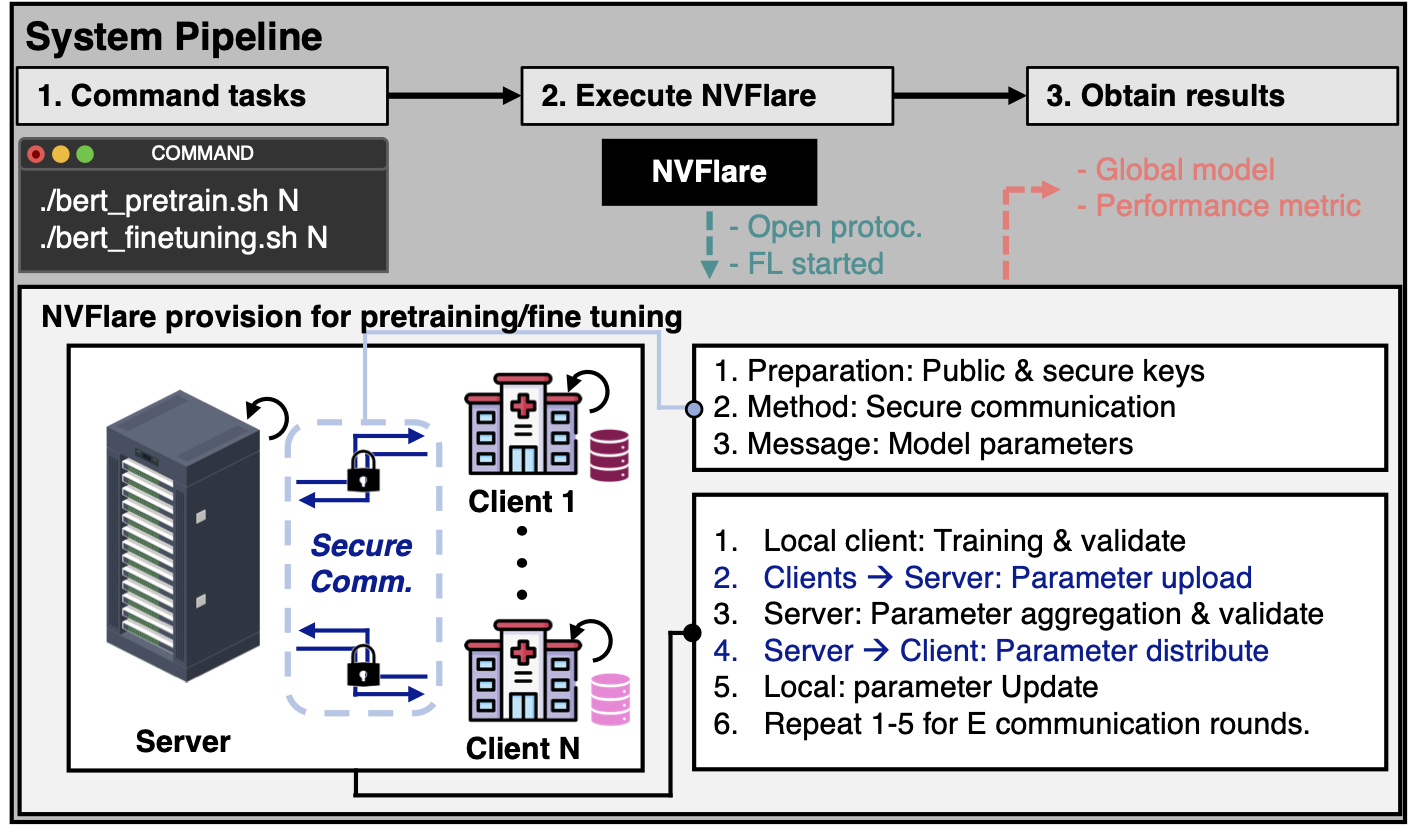}
    \caption{System pipeline.}
    \label{fig:pipeline}
    \vspace{-17pt}
\end{figure}

\subsection{NVFlare integration}
Fig.~\ref{fig:pipeline} represents the referencing system pipeline.
The system pipeline comprises three main stages: i) tasks allocation, ii) NVFlare execution, and iii) obtaining results. The task allocation stage involves the allocation of pretraining and finetuning processes. Upon executing NVFlare execution, the system generates NVFlare provision, which establishes server-client protocols, followed by federated learning. The NVFlare provision involves secure communication between the server and clients, involving a process that includes the preparation of public and secure keys, secure communication methods, and transmission of model parameters. The procedure of NVFlare operations encompasses local client training and validation, client-to-server parameter upload, server parameter aggregation and validation, server-to-client parameter distribution, and local parameter update, which repeats for $E$ communication rounds. Finally, the system obtains optimal global models and performance metrics.

\subsection{Model training}
The weights of NLP models are updated to detect patient diagnosis. To detect the patient diagnosis, this paper adopts binary classification as a task. The task is to detect patients with adverse drug reactions (ADR). We have collected electronic health records of 8,638 patients with clopidogrel prescriptions (1,824 patients were identified as treatment failure cases)~\cite{INGULEE2022}.

We also adopt a widely-used pre-training objective for transformers, namely the masked-language-model (MLM) technique, as introduced in the BERT model~\cite{devlin2018bert}. The MLM objective aims to predict the original tokens of a sentence from their masked versions to enable the model to learn to comprehend the context and semantics of the language. In this study, a masking probability of $p=0.15$ is utilized for MLM, which involves masking 15\% of the tokens in each sequence. To regulate the BERT model, 10\% of the tokens were not masked but were included in the loss calculation. During the MLM training process, the model generates probabilities for each token in the vocabulary at the masked positions, which are compared with the ground truth token for the masked positions using the cross-entropy loss. Through minimizing the loss function during training, the model learns to produce contextually accurate token predictions, thereby improving its understanding of the semantics and structure of the language.

\section{Demonstration} \label{sec:4}

\subsection{Experiment Setup}\label{sec:4-1}

To investigate the performance of the proposed framework, a feasibility testing on BERT pretraining and a comparison of three models (\textit{e.g.}, LSTM, BERT, BERT-mini) in centralized, FL, and standalone training modes are studied.
These studies are conducted with eight clients and two Linux machines. The rest of the simulation parameters are listed in Tables \ref{tab:setup} and \ref{tab:2}. 

\begin{table}[t!]
\centering
\scriptsize
\caption{Parameters used in this paper.}
\label{tab:setup}
\begin{tabular}{l|l}
\hline
\textbf{Description} &  \textbf{Values} \\ \hline
Number of clients & 8 \\\hline
\multirow{6}{*}{Hardware spec.} & $\bullet$ Machine 1 (Local server)\\
                                & \quad $*$ OS: Ubuntu 20.04 LTS, \\
                                & \quad $*$ CPU: Intel Xeon E5-2638 (2ea),\\ 
                                & \quad $*$ GPU: NVIDIA RTX 2080 Ti (4ea), \\
                                & \quad $*$ RAM: 128 GB\\
                                & $\bullet$ Machine 2 (AWS server, p3.8xlarge)\\
                                \hline
\multirow{2}{*}{Software info.} & PyTorch v11.3, CUDA v11.7, NVFlare v2.2, \\
                          & MLM-Pytorch , X-Transformers\\ 
                          \hline
\multirow{4}{*}{Data info.} & $\bullet$ \# of train data (pretraining): 453,377,\\
                            & $\bullet$ \# of valid. data (pretraining): 8,683,\\
                            & $\bullet$ \# of train data (validation): 6,927,\\
                            & $\bullet$ \# of valid. data (validation): 1,732,\\
                            \hline
\multirow{1}{*}{Learning info.} & $\bullet$ Optimizer/learning rate: Adam, $10^{-2}$ \\
                                \hline

\end{tabular}
\vspace{-6pt}
\end{table}
\begin{table}[t!]
\centering
\scriptsize
\caption{Medical NLP models used in this paper.}\label{tab:2}
\begin{tabular}{c|c|c|c}
    \hline
     Specification/Model & BERT & BERT-mini & LSTM  \\ \hline \hline
      Hidden dimension & $128$ & $50$ & $128$ \\ \hline
      \# of attention heads & $6$ & $2$ & - \\ \hline
      \# of hidden layers & $12$ & $6$ & 3 \\ \hline
\end{tabular}
    \vspace{-16pt}
\end{table}
\begin{figure}[b!]\centering
\includegraphics[width=.9\columnwidth]{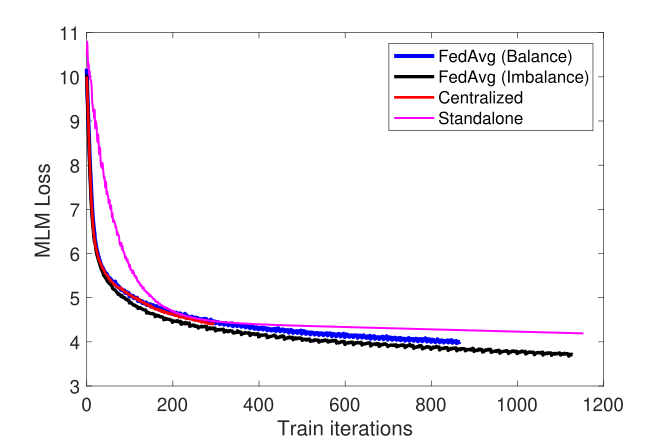}
\caption{MLM loss.}\label{fig:mlm}
    \vspace{-18pt}
\end{figure}

\subsection{Results}\label{sec:4-2}
\subsubsection{Feasibility study on BERT pretraining}  
In this feasibility study on BERT pretraining, the paper aims to investigate four distinct training schemes: \textit{1) BERT using centralized data}, \textit{2) BERT utilizing a small dataset}, \textit{3) BERT trained on imbalanced data}, and \textit{4) BERT using balanced data}. It is noteworthy that BERT employing centralized data is considered the upper bound of performance metric, while BERT utilizing a small dataset is regarded as the lower bound of performance metric. On the other hand, BERT trained on imbalanced data and balanced data represent the main focus of the FL schemes. The data imbalance is implemented by splitting the data into ratios of $\{0.29, 0.22, 0.17, 0.14, 0.09, 0.04, 0.03, 0.02\}$, with the number of data points identical for each client in the BERT using balanced data scheme.
The MLM loss is found to be comparatively lower in BERT using centralized data, BERT using balanced data, and BERT trained on imbalanced data, as well as in BERT employing a small dataset. The MLM loss begins at 10.7 and ultimately reaches 3.5 for BERT utilizing centralized data, BERT trained on imbalanced data, and BERT using balanced data. In contrast, BERT utilizing a small dataset attains only an MLM loss of 4.4, indicating that BERT with a decentralized approach is insufficient in generating medical knowledge effectively.

\begin{figure*}[t!]
    \centering
    \includegraphics[width=.87\textwidth]{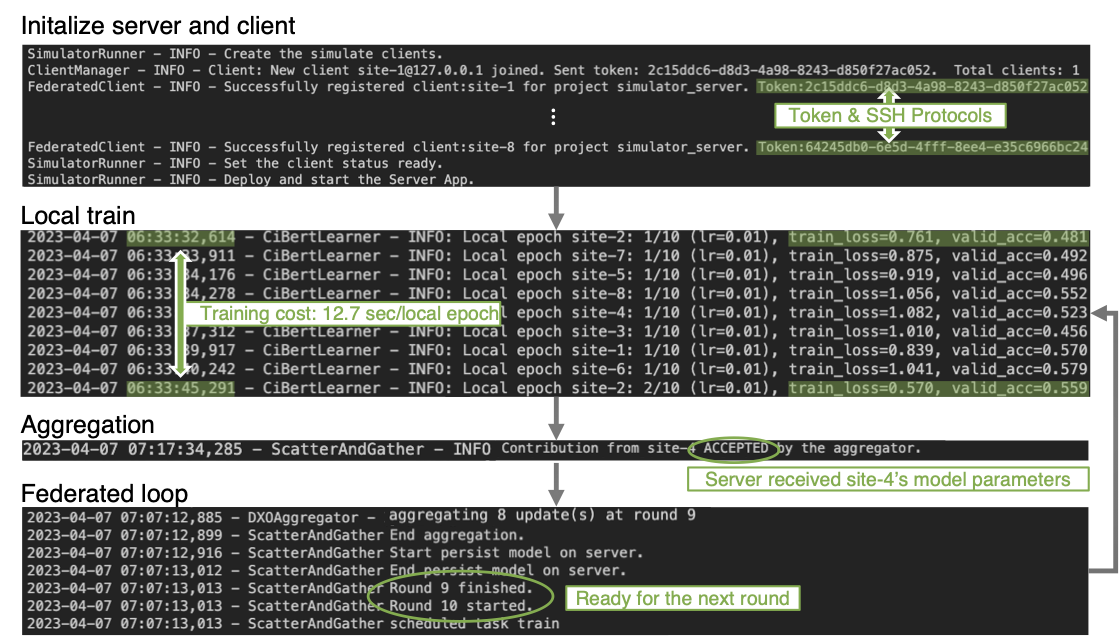}
    \caption{Demonstraion example of BERT fine-tuning.}
    \label{fig:demo}
    \vspace{-17pt}
\end{figure*}

\subsubsection{Feasibility study on recursive model in FL}
This study investigates the feasibility of FL through the utilization of three models: LSTM, BERT, and BERT-mini. 
Table~\ref{tab:top1} presents the performance of these models in centralized, FL, and standalone approaches, respectively. The LSTM-based model exhibits the highest performance, with values of ${87.9\%, 87.5\%, 67.3\%}$ for centralized, FL, and standalone approaches, respectively. Similarly, BERT and BERT-mini demonstrate similar performance tendencies across the schemes. However, the LSTM-based model outperforms both BERT and BERT-mini.

\begin{table}[t!]
\scriptsize
\caption{Top-1 accuracy [\%] of various NLP models.}\label{tab:top1}
\centering
\begin{tabular}{c|ccc}
\hline
{Schemes/Model} & BERT & BERT-mini & LSTM \\\hline \hline
 Centralized & 80.1 & 72.7 &  87.9 \\ \hline
 Standalone  & 72.2 & 68.5 &  67.3 \\ \hline
 FL          & 80.1 & 72.3 &  87.5 \\\hline
\end{tabular}
    
\end{table}

\subsubsection{Limitation on BERT} 
Although BERT is generally recognized for its superior performance in comparison to LSTM across a wide range of natural language processing tasks, there are certain circumstances in which LSTM may outperform BERT. These situations can be attributed to the following factors:
Firstly, task characteristics play a significant role, as LSTM may possess a more appropriate structure for specific tasks relative to BERT. For example, LSTM may excel in sequence modeling or time series data processing tasks. Secondly, dataset size is a crucial determinant, since LSTM can be effectively trained with relatively smaller amounts of data. Consequently, in scenarios with limited datasets, LSTM may surpass BERT's performance.
Furthermore, overfitting is an issue that plagues large models like BERT, making them susceptible to poor generalization when applied to small datasets. In contrast, LSTM can achieve superior generalization due to its fewer parameters. Finally, the performance disparity may emerge from differences in optimization methods employed during model training, including hyperparameter settings, learning rate, loss functions, and others. Under certain conditions, LSTM may learn more efficiently than BERT.

\subsubsection{Demonstrations} Fig.~\ref{fig:demo} illustrates the implementation of the NVFlare integrated framework, showcasing a BERT fine-tuning example. Initially, the server and clients are established, with token and SSH-based protocols employed to ensure secure communication channels. Subsequently, each client proceeds to train its model locally, taking an average of 12.7 seconds per local epoch to complete the process. Upon completion, the aggregator gathers the local model parameters from each client. Once all model parameters have been collected, the subsequent federated round commences.

\section{Conclusion and Future Research Directions}\label{sec:5}
This paper proposes a practical approach to medical NLP by integrating FL with NVFlare. The proposed framework addresses the challenges of data privacy and regulatory compliance while maintaining high accuracy and performance. Furthermore, this study compares the performance of LSTM-based models to BERT and BERT-mini across various training settings: centralized, FL, and standalone. While LSTM outperforms BERT in the experiments, this outcome may be due to task characteristics, dataset size, overfitting, or optimization-related factors.

Future research directions includes investigating the impact of different tasks and dataset sizes on the performance of LSTM and BERT in medical NLP applications.

\bibliographystyle{IEEEtran}
\bibliography{reference}
\end{document}